\documentclass[11pt]{article}

\usepackage[margin=1in]{geometry}
\usepackage{amsmath,amssymb}
\usepackage{graphicx}
\usepackage{hyperref}
\usepackage{algorithm}
\usepackage{algpseudocode}
\usepackage{booktabs}

\title{Cubic Spline Interpolation for Smooth Trajectory Generation \\ in Autonomous Motion Systems}

\author{
Aarush Mandayam$^{1}$ \quad Sriram Aditya Gundla$^{1}$ \quad Elliott Tammaro, PhD$^{1,2}$ \\[4pt]
\small $^{1}$10x Accel \\
\small $^{2}$Chestnut Hill College, Philadelphia, Pennsylvania
}

\date{\today}

\begin{document}
\maketitle

\begin{abstract}
Trajectories for robots, drones, and other autonomous vehicles must pass through a sequence of prescribed waypoints while remaining physically realizable, meaning the underlying position signal should have continuous velocity and acceleration. Piecewise-linear interpolation between waypoints violates this requirement: velocity changes discontinuously at each waypoint, implying infinite acceleration. We describe a program that fits piecewise cubic splines to a set of waypoints specified as position-time pairs $(x_i, t_i)$, enforcing continuity of the function value, first derivative, and second derivative across segment boundaries. The spline coefficients are obtained by assembling the interpolation and continuity conditions into a single linear system and solving for the unknown coefficients directly, removing the need to derive the per-segment equations by hand. We demonstrate the approach on two illustrative tasks: generating smooth free-form paths from a sparse set of user-specified points, and reconstructing continuous digital traces of handwritten signatures from sampled pen positions. We discuss the computational cost of the direct linear-system approach relative to the classical tridiagonal (moment-based) formulation, and outline the extensions needed to apply this method to multi-axis robot and drone trajectories subject to velocity and acceleration limits.
\end{abstract}

\noindent\textbf{Keywords:} cubic splines, trajectory generation, motion planning, smooth interpolation, autonomous vehicles

\section{Introduction}

A recurring problem in robotics, computer animation, and autonomous vehicle control is generating a trajectory that passes through a specified sequence of waypoints while producing motion that is physically executable. A naive solution connects consecutive waypoints with straight line segments (piecewise linear interpolation). This is simple to compute, but it is unsuitable for actuated physical systems: at each interior waypoint, the direction and often the magnitude of the velocity vector changes instantaneously. Since acceleration is the time derivative of velocity, a discontinuous velocity implies an infinite acceleration at that instant, which no real actuator, motor, or propulsion system can produce.

This limitation motivates the use of smoother interpolants. A trajectory that is $C^1$ continuous (continuous position and velocity) removes the infinite-acceleration problem at the position level, but a trajectory that is $C^2$ continuous (continuous position, velocity, and acceleration) is preferable for most physical systems, since acceleration commands are what feedback controllers on robots and drones ultimately track. Piecewise cubic polynomials are the lowest-degree polynomials that can be joined together to form a globally $C^2$ curve while still leaving enough free parameters to interpolate an arbitrary sequence of waypoints; this is the classical cubic spline \cite{deboor2001}.

In more demanding aerial robotics applications, still higher continuity (up to the fourth derivative, or ``snap'') is used so that the commanded thrust and body rates vary smoothly, as in minimum-snap trajectory generation for quadrotors \cite{mellinger2011}. Cubic splines represent the simplest member of this family: they guarantee only $C^2$ continuity, but they are correspondingly cheaper to compute and are sufficient for many ground robot, camera path, and animation applications where jerk and higher derivatives are not explicitly constrained.

In this work we describe a program that automates the construction and solution of the cubic spline interpolation problem for a set of waypoints given as $(x_i, t_i)$ pairs. Rather than deriving the spline coefficients for each new set of waypoints by hand, the program assembles the full linear system of interpolation and continuity constraints, and solves for the coefficient vector directly. We apply the resulting tool to two tasks that are not typically framed as robotics problems, but which exercise exactly the same underlying interpolation machinery: (1) generating smooth free-form curves (``abstract drawings'') from a small set of user-supplied points, and (2) reconstructing a smooth digital trace of a handwritten signature from a sparse set of sampled points lying on the signature. Both tasks illustrate that once a program can turn an arbitrary list of waypoints into a $C^2$ curve, the same machinery is directly reusable for generating feasible motion commands for physical autonomous systems.

\section{Related Work}

Cubic spline interpolation is a classical topic in numerical analysis; de Boor's text \cite{deboor2001} remains the standard reference for both the theory (existence, uniqueness, and optimality properties of the interpolating spline) and the numerical construction of splines via divided differences and B-spline recursions. The formulation used in this paper follows the direct polynomial-coefficient construction described there, rather than the B-spline basis.

In robotics and aerial vehicle control, piecewise polynomial trajectories are widely used to convert a sequence of waypoints into a reference signal that a lower-level controller can track. Mellinger and Kumar \cite{mellinger2011} generate trajectories for quadrotors by minimizing the integral of the square of the fourth derivative (snap) of position, subject to waypoint and boundary constraints, using piecewise polynomials of higher degree than cubic. Their formulation illustrates the general pattern followed here: waypoint interpolation constraints plus derivative-continuity constraints at segment boundaries, assembled into a linear system whose solution gives the polynomial coefficients. Cubic splines correspond to the special case where only continuity up to the second derivative is required, which reduces the polynomial degree needed per segment from the degree required for minimum-snap trajectories, at the cost of not directly controlling jerk or snap.

Cubic splines and related low-order piecewise-polynomial curves have also been used outside of vehicle motion planning for exactly the same reason they are useful here: they turn a sparse set of sample points into a smooth, evaluable curve. In signature processing, cubic splines have been used to smooth the raw, jagged pen-position sequences captured by a digitizing tablet before further feature extraction \cite{saffar2018}, and Bézier variants of piecewise cubics have been used to vectorize and reconstruct handwritten signature contours. This is directly analogous to the signature-digitization application described in Section~\ref{sec:applications} of this paper, where a small number of sampled points on a signature are used to reconstruct a continuous curve.

\section{Problem Formulation}

We are given a set of $n+1$ waypoints $(x_0, t_0), (x_1, t_1), \dots, (x_n, t_n)$, with $t_0 < t_1 < \dots < t_n$, where $t_i$ plays the role of the interpolation parameter (typically time, but the same formulation applies to any monotonically increasing parameterization such as arc length). The goal is to construct a function $S(t)$ that:

\begin{enumerate}
    \item passes through every waypoint, i.e., $S(t_i) = x_i$ for all $i$;
    \item is twice continuously differentiable on $[t_0, t_n]$, i.e., $S \in C^2[t_0, t_n]$.
\end{enumerate}

The cubic spline construction represents $S(t)$ as $n$ piecewise cubic polynomials, one per interval $[t_i, t_{i+1}]$:
\begin{equation}
    S_i(t) = a_i + b_i (t - t_i) + c_i (t - t_i)^2 + d_i (t - t_i)^3, \qquad t \in [t_i, t_{i+1}], \quad i = 0, \dots, n-1.
    \label{eq:segment}
\end{equation}

This introduces $4n$ unknown coefficients $\{a_i, b_i, c_i, d_i\}_{i=0}^{n-1}$. These are determined by the following conditions:

\textbf{Interpolation conditions} ($2n$ equations). Each segment must match the waypoint values at both of its endpoints:
\begin{align}
    S_i(t_i) &= x_i, & i = 0, \dots, n-1, \\
    S_i(t_{i+1}) &= x_{i+1}, & i = 0, \dots, n-1.
\end{align}

\textbf{First-derivative continuity} ($n-1$ equations). At each interior waypoint, the velocity computed from the segment on the left must match the velocity computed from the segment on the right:
\begin{equation}
    S_i'(t_{i+1}) = S_{i+1}'(t_{i+1}), \qquad i = 0, \dots, n-2.
\end{equation}

\textbf{Second-derivative continuity} ($n-1$ equations). Similarly, acceleration must match across each interior waypoint:
\begin{equation}
    S_i''(t_{i+1}) = S_{i+1}''(t_{i+1}), \qquad i = 0, \dots, n-2.
\end{equation}

These conditions total $2n + 2(n-1) = 4n - 2$ equations, two short of the $4n$ unknowns. The remaining two degrees of freedom are fixed by a choice of boundary condition at the two ends of the trajectory. Common choices include:

\begin{itemize}
    \item \emph{Natural boundary conditions}: $S_0''(t_0) = 0$ and $S_{n-1}''(t_n) = 0$, which impose zero curvature (and, in a trajectory-tracking context, zero commanded acceleration) at the start and end of motion.
    \item \emph{Clamped boundary conditions}: $S_0'(t_0)$ and $S_{n-1}'(t_n)$ are set to specified values, useful when a known initial or final velocity is required (e.g., a robot starting or ending at rest, or matching the velocity of an adjoining trajectory segment).
\end{itemize}

\subsection{Matrix Formulation}

All $4n - 2$ continuity and interpolation equations, together with the two boundary equations, are linear in the unknown coefficients. They can therefore be written as a single matrix equation
\begin{equation}
    A \mathbf{c} = \mathbf{b},
    \label{eq:linear-system}
\end{equation}
where $\mathbf{c} \in \mathbb{R}^{4n}$ stacks the unknowns $\{a_i, b_i, c_i, d_i\}$, $A \in \mathbb{R}^{4n \times 4n}$ is a sparse matrix built entirely from the waypoint times $t_i$ (evaluating the appropriate powers of $(t_{i+1} - t_i)$ at each row), and $\mathbf{b} \in \mathbb{R}^{4n}$ collects the waypoint values $x_i$ and the boundary-condition constants, with zeros elsewhere. Once $A$ and $\mathbf{b}$ are assembled, the coefficient vector is recovered as
\begin{equation}
    \mathbf{c} = A^{-1} \mathbf{b}.
\end{equation}

This is the approach implemented in our program: rather than deriving the entries of $A$ and $\mathbf{b}$ by hand for each new set of waypoints, the program constructs $A$ and $\mathbf{b}$ programmatically from the input list of $(x_i, t_i)$ pairs and computes the matrix inverse (or, equivalently, solves the linear system) to obtain all spline coefficients in one step.

\section{Implementation}

The program accepts a list of waypoints as $(x, t)$ pairs. For $n+1$ waypoints, it constructs the $4n \times 4n$ matrix $A$ and right-hand-side vector $\mathbf{b}$ described in Section~3.1 by iterating over each segment and each continuity condition, then solves Equation~\ref{eq:linear-system} for the coefficient vector $\mathbf{c}$. The resulting per-segment cubic coefficients are stored and can be evaluated at any query time $t \in [t_0, t_n]$ by locating the containing segment and evaluating Equation~\ref{eq:segment}.

Algorithm~\ref{alg:spline} summarizes the procedure.

\begin{algorithm}[h]
\caption{Cubic spline fitting via direct linear system solution}
\label{alg:spline}
\begin{algorithmic}[1]
\Require Waypoints $(x_0, t_0), \dots, (x_n, t_n)$; boundary condition type
\State Initialize $A \gets \mathbf{0}^{4n \times 4n}$, $\mathbf{b} \gets \mathbf{0}^{4n}$
\For{$i = 0$ to $n-1$}
    \State Fill rows of $A, \mathbf{b}$ for $S_i(t_i) = x_i$ and $S_i(t_{i+1}) = x_{i+1}$
\EndFor
\For{$i = 0$ to $n-2$}
    \State Fill rows of $A$ for $S_i'(t_{i+1}) = S_{i+1}'(t_{i+1})$
    \State Fill rows of $A$ for $S_i''(t_{i+1}) = S_{i+1}''(t_{i+1})$
\EndFor
\State Fill the two remaining rows of $A, \mathbf{b}$ according to the chosen boundary condition
\State $\mathbf{c} \gets A^{-1}\mathbf{b}$ \Comment{solve the assembled linear system}
\State \Return per-segment coefficients extracted from $\mathbf{c}$
\end{algorithmic}
\end{algorithm}

\textbf{Computational note.} Solving the full $4n \times 4n$ system by direct matrix inversion costs $O(n^3)$ arithmetic operations in general. The classical alternative formulation of the cubic spline problem reduces the unknowns to the $n+1$ second derivatives (moments) at the waypoints, which produces a tridiagonal system solvable in $O(n)$ time via the Thomas algorithm \cite{deboor2001}. For the waypoint counts used in this work the $O(n^3)$ cost of direct inversion is not a practical bottleneck, but the tridiagonal formulation would be preferable for applications with large numbers of waypoints or real-time replanning requirements, such as online trajectory replanning for a drone reacting to a dynamic environment.

\section{Applications}
\label{sec:applications}

We demonstrate the spline-fitting program on two tasks that require converting a sparse set of sample points into a smooth, continuous curve.

\subsection{Free-form path synthesis}

Given a small number of $(x, t)$ waypoints placed by hand, the program produces a smooth curve passing through all of them.

\subsection{Signature digitization}

A handwritten signature can be represented as a sequence of pen positions sampled over time as the signature is written. Treating these samples as waypoints, the program fits a cubic spline to reconstruct a continuous digital trace of the signature from a reduced set of points. Cubic splines have previously been used in the signature-processing literature specifically to smooth raw, noisy pen-position sequences prior to further analysis \cite{saffar2018}; the present application uses the same underlying interpolation for a related purpose, converting a sparse point set into a continuous curve suitable for digital storage or display. 

\section{Discussion and Limitations}

The formulation presented here treats each spatial axis independently: a waypoint sequence $(x_i, t_i)$ produces a scalar function $x(t)$. Extending this to multi-dimensional trajectories, as required for planar or 3D robot and drone motion, is done by fitting an independent cubic spline to each coordinate axis ($x(t)$, $y(t)$, and, for aerial vehicles, $z(t)$) using the same set of waypoint times $t_i$. Because all axes share the same knot sequence, the resulting vector-valued trajectory $\mathbf{r}(t) = (x(t), y(t), z(t))$ is $C^2$ in each component and therefore has well-defined, continuous velocity and acceleration vectors.

Two limitations are worth noting. First, the cubic spline construction used here enforces only $C^2$ continuity and does not explicitly bound the magnitude of velocity or acceleration, nor does it constrain jerk; a trajectory that is technically feasible in the continuity sense may still command accelerations that exceed what a given robot or drone's actuators can produce. Approaches such as minimum-snap trajectory generation \cite{mellinger2011} address this by directly optimizing a cost that penalizes higher derivatives and by incorporating actuator limits into the optimization, at the cost of a higher-degree polynomial and a more expensive optimization problem than a direct linear solve. Second, the assignment of times $t_i$ to waypoints in this work is taken as given; in a full trajectory-planning pipeline, choosing an appropriate time allocation across waypoints (so that, for example, closely spaced waypoints are not required to be traversed unrealistically quickly) is itself a design decision that affects the feasibility of the resulting trajectory.

\section{Conclusion}

We presented a program that automates the construction and solution of the cubic spline interpolation problem for a sequence of position-time waypoints, producing trajectories that are continuous in position, velocity, and acceleration and therefore avoid the infinite-acceleration problem inherent to piecewise-linear waypoint following. By assembling the interpolation and continuity conditions into a single linear system and solving directly for the spline coefficients, the program removes the need to manually derive per-segment equations for each new waypoint set. We illustrated the approach on free-form curve drawing and signature digitization, two tasks that use the same underlying machinery required for smooth robot and drone trajectory generation. Future work includes extending the implementation to multi-axis trajectories with explicit velocity and acceleration bounds, adopting the more computationally efficient tridiagonal solution method for larger waypoint sets, and evaluating the approach quantitatively against alternative trajectory generation methods.

\end{document}